\documentclass[times, review, 10pt]{elsarticle}
\usepackage{amssymb}
\usepackage{amsmath}
\usepackage{multirow}
\usepackage{booktabs}
\usepackage{hyperref}
\usepackage{tabularx}
\usepackage{algorithmic}
\usepackage{algorithm}
\usepackage{array}
\usepackage{xcolor} % 引入颜色包

\journal{Pattern Recognition}

\begin{document}

\begin{frontmatter}


%% use the tnotetext command for theassociated footnote;Graphical Abstract

\title{Unsupervised Domain Adaptation via Style-Aware Self-intermediate Domain}

\author{Lianyu~Wang\corref{cor1}\fnref{label1}}
\author{Meng~Wang\corref{cor1}\fnref{label2}}
\author{Daoqiang~Zhang\corref{cor2}\fnref{label1}}
\author{Huazhu~Fu\corref{cor2}\fnref{label2}}

\affiliation[label1]{organization={Key Laboratory of Brain-Machine Intelligence Technology, Ministry of Education},
            city={Nanjing},
            postcode={211106},
            country={China}}

\affiliation[label2]{organization={Institute of High Performance Computing, A*STAR},
            postcode={138632},
            country={Singapore}}
\cortext[cor1]{Lianyu Wang and Meng Wang were contributed equally to this work.}
\cortext[cor2]{Daoqiang Zhang and Huazhu Fu are co-corresponding authors.}


%% Abstract
\begin{abstract}
% Unsupervised domain adaptation (UDA) has attracted considerable attention, which transfers knowledge from a label-rich source domain to a related but unlabeled target domain. Reducing inter-domain differences has always been a crucial factor in improving performance in UDA, especially for tasks where there is a large gap between source and target domains. To this end, we propose a novel Style-aware Self-Intermediate Domain (SSID) to bridge the large domain gap and transfer knowledge while alleviating the loss of class-discriminative information. 
Unsupervised Domain Adaptation (UDA) has garnered significant attention for its ability to transfer knowledge from a label-rich source domain to a related but unlabeled target domain, with minimizing inter-domain discrepancies being crucial, especially when a substantial gap exists between the domains. To address this, we introduce the novel Style-aware Self-Intermediate Domain (SSID), which effectively bridges large domain gaps by facilitating knowledge transfer while preserving class-discriminative information. Inspired by human transitive inference and learning capabilities, SSID connects seemingly unrelated concepts through a sequence of intermediate, auxiliary synthesized concepts. Meanwhile, an external memory bank is designed to store and update designated labeled features, ensuring the stability of class-specific and class-wise style features. Additionally, we also proposed a novel intra- and inter-domain loss functions that enhance class recognition and feature compatibility, with their convergence rigorously validated through a novel analytical approach. Comprehensive experiments demonstrate that SSID achieves accuracies of 85.4\% and 85.3\% on two widely recognized UDA benchmarks, outperforming the second-best methods by 0.94\% and 1.17\%, respectively. As a plug-and-play solution, SSID integrates seamlessly with various backbone networks, showcasing its effectiveness and versatility in domain adaptation scenarios.

\end{abstract}

% %Graphical abstract

% %Research highlights
% \begin{highlights}
% \item We propose Style-aware Self-Intermediate Domain (SSID), a novel UDA solution, to bridge domain gaps and transfer knowledge while preserving class-discriminative information. Additionally, SSID is designed to generate labeled and style-rich auxiliary features, narrowing the gap between domains.
% \item An external memory bank stores specified features based on sample indices, allowing for the extraction of stable class features and class-wise style features.
% \item We develop novel intra- and inter-domain losses based on the memory bank to enhance the network's class recognition ability and feature compatibility, respectively. Additionally, we simulate the rich latent feature space of SSID through infinite sampling and demonstrate, for the first time, the convergence of the proposed loss function.
% \item Extensive experiments on commonly used domain adaptive benchmarks validate the proposed SSID, demonstrating its effectiveness as a plug-and-play technique applicable to various UDA backbones.

% \end{highlights}

%% Keywords

\end{frontmatter}



\section{Introduction}
Deep neural networks have excelled in diverse machine learning areas, like image classification~\cite{ref1}, semantic segmentation~\cite{ref2}, and object recognition. However, these advancements heavily rely on large-scale datasets and assume consistent feature distributions between training and testing data. In reality, data collected from various environments may differ in lighting, color saturation, and hue, leading to domain gaps. Domain adaptation aims to align data distributions from different domains by mapping them into a common feature space and optimizing model parameters to minimize the discrepancy between features from different domains.
Domain adaptation can be categorized into supervised, semi-supervised, and unsupervised domain adaptation (UDA) based on the availability of labels in the target domain~\cite{ref17}. Labeling data is often time-consuming, labor-intensive, and costly, which limits performance improvement. Unlike supervised and semi-supervised approaches, UDA does not rely on labeled target domain data, making it applicable in various scenarios and deserving further attention. This paper focuses on exploring UDA methods to transfer knowledge from a labeled source domain to a different unlabeled target domain.

\begin{figure}[!t]
\centering
\includegraphics[width=0.6\linewidth,trim=310 150 300 150,clip]{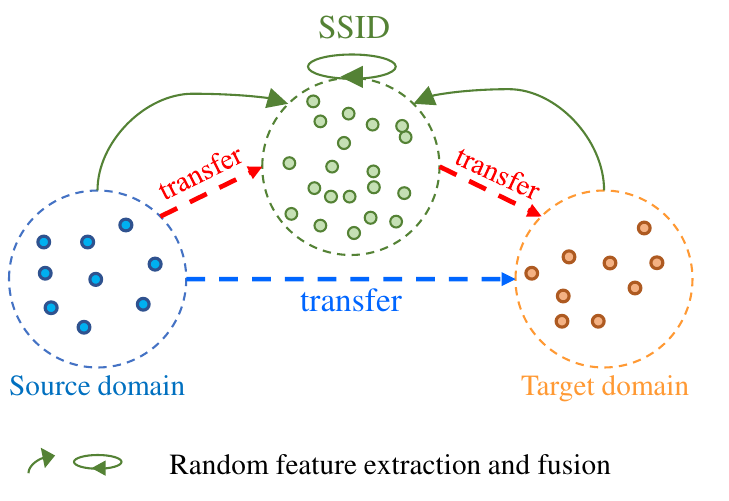}
\caption{The illustration of the proposed SSID. In the UDA task, the significant gap between the source and target domains (represented by the blue dashed line) presents a challenge for transferring class-discriminative information. The proposed SSID learning strategy addresses this by randomly selecting samples from multiple domains as anchors and computing their object and style features. It then generates a labeled intermediate auxiliary class-discriminative information by randomly combining the object and style features (indicated by the red dashed line). The labeled and style-rich SSID serves as a bridge, implicitly transferring class information from the source to the target domains.}
\label{fig_1}
\end{figure}

In recent years, various UDA methods have been developed, broadly categorized into two types: 1) difference-based methods, which explicitly reduce inter-domain differences by minimizing statistical metrics like maximum mean discrepancy, margin disparity discrepancy, and optimal transport distance; and 2) adversarial-based methods, which use a discriminator to learn domain-invariant representations adversarially by obfuscating domain-specific information. While these methods aim to improve feature transferability between domains, they face challenges. Difference-based methods may reduce class discriminability, while adversarial-based methods may overlook class-variant representations, both leading to decreased discriminability. This decline is inevitable in UDA, where the target domain lacks labels and significant domain gaps exist, causing class-discriminative information loss during domain alignment. This bias towards the source domain can degrade discriminability in the target domain~\cite{ref18}. Our work addresses these challenges by designing a robust solution to enhance feature transferability and discriminability, thereby reducing the domain gap.

Drawing inspiration from human transitive inference and learning abilities~\cite{ref21}, we propose a novel domain adaptation approach: Style-aware Self-Intermediate Domain (SSID). This method links two seemingly unrelated domains through a series of intermediate auxiliary synthesis concepts. Specifically, the source and target domains are primarily distinguished by their image styles, including lighting conditions, color saturation, hue, visual angle, and background environment, etc. When the styles of the same class vary significantly, incorrect feature alignment can occur, compromising the discriminability of the target domain. To address this, we propose generating labeled and style-rich auxiliary intermediate features to bridge the source and target domains. In SSID, we randomly extract features from multiple domains as anchors, compute their object and style features, and synthesize labeled auxiliary class-discriminative information by randomly combining these features. This labeled and style-rich SSID serves as a bridge, implicitly transferring domain information from the source to the target domain, as illustrated in Fig.~\ref{fig_1}.
Furthermore, we incorporate an external memory bank with six memory cells to store specific features of the SSID and the source domain. These feature centers are calculated and stored using labels from the source domain data to acquire stable class features and class-wise style features.
Leveraging this memory bank, we devise novel intra- and inter-domain losses for SSID to enhance class recognition ability and feature compatibility, respectively. We simulate the expansive latent feature space of SSID through infinite sampling, derive the upper bound loss function, and demonstrate the convergence of our proposed approach.

In general, we highlight our four-fold contributions:
\begin{itemize}
\item{We propose Style-aware Self-Intermediate Domain (SSID), a novel UDA solution, to bridge domain gaps and transfer knowledge while preserving class-discriminative information. Additionally, SSID is designed to generate labeled and style-rich auxiliary features, narrowing the gap between domains.}
\item{An external memory bank stores specified features based on sample indices, allowing for the extraction of stable class features and class-wise style features.}
\item{We develop novel intra- and inter-domain losses based on the memory bank to enhance the network's class recognition ability and feature compatibility, respectively. Additionally, we simulate the rich latent feature space of SSID through infinite sampling and demonstrate, for the first time, the convergence of the proposed loss function.}
\item{Extensive experiments on commonly used domain adaptive benchmarks validate the proposed SSID, demonstrating its effectiveness as a plug-and-play technique applicable to various UDA backbones.} 
\footnote{The code will be released on \url{https://github.com/LyWang12/SSID}}
\end{itemize}

\section{Related Work}
UDA has garnered attention in the deep learning community for minimizing distribution differences between different but related domains. Many methods have been proposed and applied in cross-domain applications. These methods can be broadly categorized into two types: difference-based UDA and adversarial-based UDA.

\subsection{Difference-Based UDA Method}
Difference-based UDA alleviates domain differences by minimizing statistical differences. Long~\textit{et al.}~\cite{ref6} minimized the maximum mean difference for task-specific layers to narrow the domain gap explicitly. Based on the optimal transfer distance, Zhang~\textit{et al.}~\cite{ref13} designed optimal transfer models to perform feature alignment in domains. In addition to these earlier efforts, regularization techniques, like entropy constraints or maximum prediction rank~\cite{ref15}, can implicitly constrain the cross-domain feature space. 
While these methods help align similar features between domains, the lack of labels in the target domain can lead to incorrect distance calculations for similar but different-class features, reducing discriminability. 

\subsection{Adversarial-Based UDA Method}
Adversarial-based UDA introduces a discriminator to adversarially learn domain-invariant representations by obfuscating domain-specific information. Methods like CDAN~\cite{ref17} employ a gradient inversion layer and a domain discriminator to confuse the feature distributions of the two domains. To preserve inter-class discriminability while obscuring domain features, Saito~\textit{et al.}~\cite{ref19} proposed a novel adversarial paradigm where the adversarial process occurs between the feature extractor and the classifier. 
These methods treat features within a domain, compelling the feature extractor to learn domain-invariant representations by confusing the domain discriminator. However, this approach may confuse finer intra-domain information, such as class-specific features, thereby reducing discriminability.

\begin{figure*}[!t]
\centering
\includegraphics[width=1\linewidth,trim=45 30 45 40,clip]{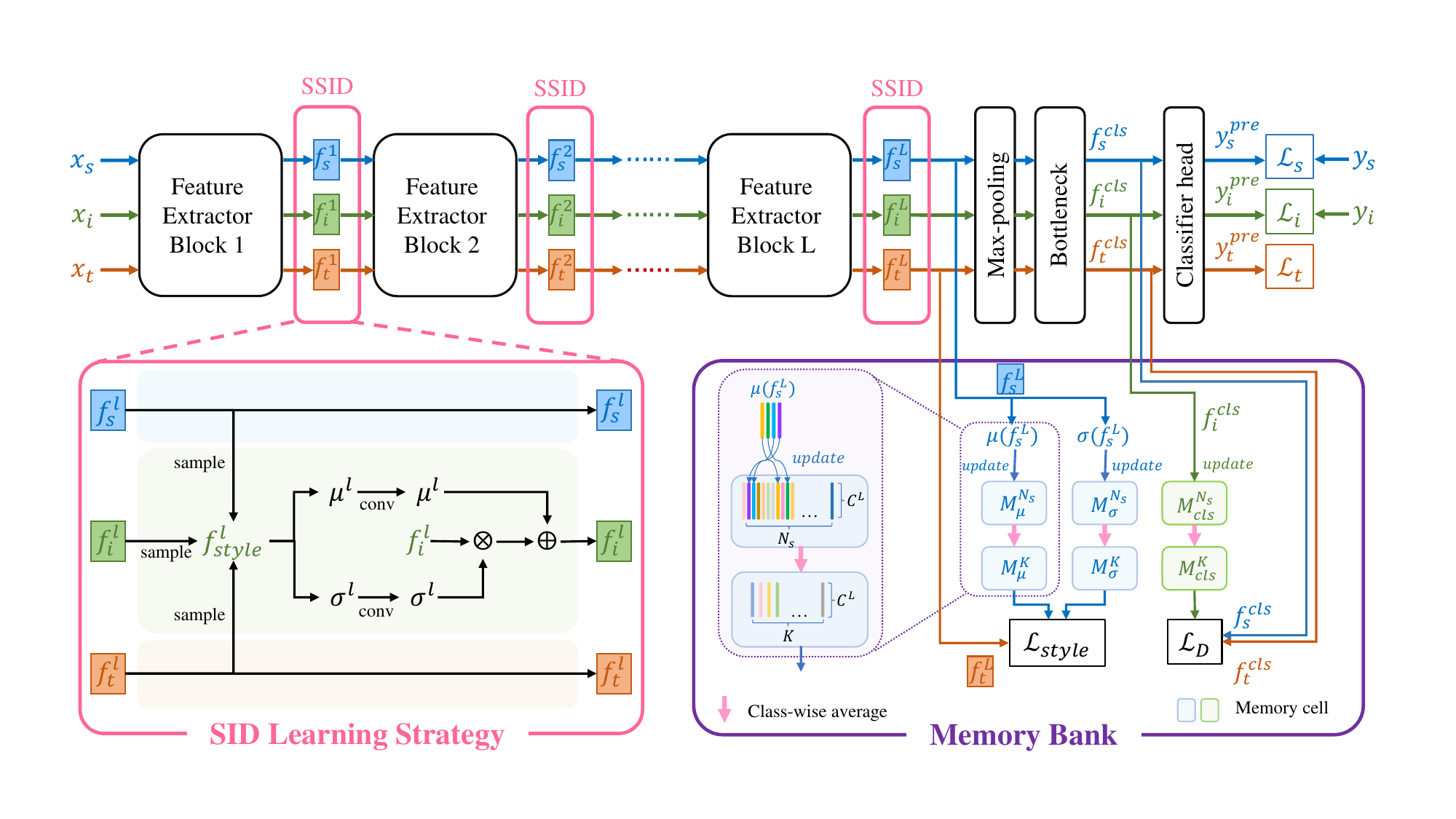}
\caption{The illustration of our proposed framework, including the SSID learning strategy, the external memory bank, and intra/inter-domain losses. Samples from the source domain, SSID, and target domain are fed into the feature extractor in parallel, denoted by blue, green, and orange, respectively. The SSID learning strategy is deployed after each feature extractor block, followed by the external memory bank.}
\label{fig_2}
\end{figure*} 

\subsection{Other UDA Method}
Some studies tried to embed source and target data into a Grassmann manifold and learn a specific geodesic path between them, but these methods are not easily applicable to deep models~\cite{ref31}. Following them, Dai~\textit{et al.}~\cite{ref32} proposed a person re-identification method based on the shortest geodesic path definition. However, their method is designed for supervised domain adaptation, as their bridging loss involves labels from the target-train set. Sun~\textit{et al.}~\cite{ref33} proposed a GAN-based method to generate target domain-like images from the source domain, but it is computationally expensive. Li~\textit{et al.}~\cite{ref34} introduced a transferable semantic augmentation approach to enhance classifier adaptation ability but overlooked the feature mixing association between different categories.

Recently, Transformer-based UDA approaches have been explored. TransDA~\cite{ref36} focuses on source-free domain adaptation, introducing a novel self-supervised knowledge distillation approach with target pseudo-labels. However, it emphasizes object features, potentially weakening useful style features. TVT~\cite{ref37} enhances feature diversity and separation using class tokens and discriminative clustering, which are often compromised during adversarial domain alignment. However, it has a large number of parameters and is only suitable for Transformer-based networks.

Our method differs by considering transfer from the source to the target domain, generating a labeled, style-rich SSID as a bridge through random fusion to implicitly transfer class-discriminative information, improving transferability and discriminability simultaneously.

\section{Methodology}
In this section, we elaborate on each component of our proposed SSID, which aims to enhance the transferability of cross-domain features and leverage class-discriminative information. This approach generates labeled cross-domain representations that implicitly preserve class-discriminative information, thereby narrowing the domain gap in UDA.
Fig.~\ref{fig_2} illustrates our proposed framework. \(x_{s}\), \(x_{i}\) and \(x_{t}\) are samples from source domain \(S = \left\{ {\left( {{x_{sm}},{y_{sm}}} \right)} \right\}_{m = 1}^{{N_s}}\), SSID \(I = \left\{ {\left( {{x_{in}},{y_{in}}} \right)} \right\}_{n = 1}^{{N_s}}\) and target domain \(T = \left\{ {\left( {{x_{tp}}} \right)} \right\}_{p = 1}^{{N_t}}\) respectively, which are fed into the feature extractor in parallel, with \(I\) initialized by \(S\), where \(y \in \left\{ {1,2, \ldots ,K} \right\}\) is the label of \(x\), \(N\) and \(K\) denote the number of sample and class respectively. 
The Vision Transformer (ViT)~\cite{ref38} is used as the feature extractor to extract more general representations as its outstanding performance in many computer vision tasks. 
Our feature extractor comprises \(L\) feature extractor blocks followed by the SSID learning strategy detailed in Section~\ref{subsec31}. This is followed by the max-pooling layer, the bottleneck layer, and the classifier head layer. The specified output of the feature extractor and bottleneck layer are stored and updated in the external memory bank (Section~\ref{subsec32}) for subsequent loss function computation (Section~\ref{subsec33}).

\begin{figure*}[!t]
\centering
\includegraphics[width=1\linewidth,trim=90 95 90 95,clip]{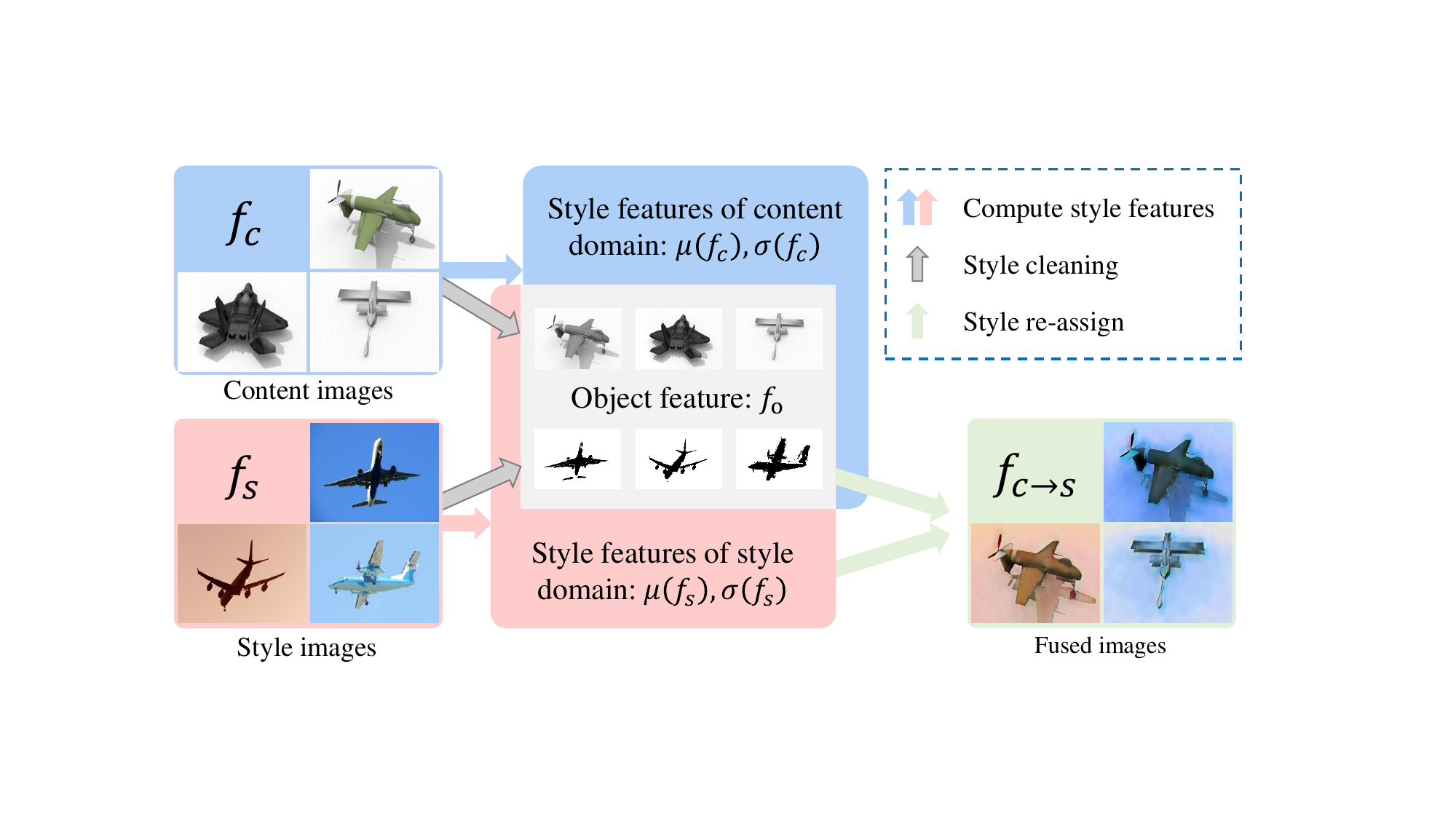}
\caption{The illustration of style fusion from content images to style images. Style features of the style images are used to perform style re-assign on the object features of the content images to obtain the fused features. The content and style of the fused images are consistent with the content images and the style images, respectively. Features can be visualized by a trained generator.}
\label{fig_3}
\end{figure*}

\subsection{Style-aware Self-Intermediate Domain (SSID) Learning Strategy}
\label{subsec31}

Style refers to weakly semantically related cues in the feature map, such as lighting conditions, hue, and color saturation. Samples from different domains exhibit large stylistic differences, making UDA tasks challenging. Previous work focused on improving feature transferability by suppressing weak style-related features, but this can lead to loss of fine class-discriminative features. Instead, we enhance model robustness by enriching cross-domain style information.

Style transfer assumes styles are homogeneous, with images considered similar if their extracted features share the same statistics, such as mean and variance~\cite{ref42}. Object features \(f_o\), containing only semantic information, can be derived by removing style features, as:
\begin{equation}
\label{eq1}
f_o = \frac{{f_c - {\mu(f_c)}}}{{{\sigma(f_c)}}},
\end{equation} 
where \(\mu(f_c)\) and \(\sigma(f_c)\) denote the mean and variance of extracted feature \(f_c\). Furthermore, style can be re-assign by \(f_o \cdot \gamma  + \beta\), where \(\gamma\) and \(\beta\) are learned parameters. Afterward,~\cite{ref42} further explored adapting \(f_o\) to arbitrarily given style by using style features of another extracted feature instead of learned parameters. 
Building on these ideas, we propose a feature fusion method, depicted in Fig.~\ref{fig_3}, for the UDA task. First, input images are divided into content images and style images, with their features extracted by a trained classifier denoted as \(f_c\) and \(f_s\), respectively. Next, the style features \(\mu (f_c)\) and \(\sigma (f_c)\) of content images are isolated to derive object features \(f_o\). Thirdly, the styles of style images are transferred to  \(f_o\) to obtain fused features \(f_{c \rightarrow s}\). This process ensures that the content and style of 
\(f_{c \rightarrow s}\) align with the content and style images, respectively. The feature fusion method aims to enrich the distribution of a class in the feature space by generating another style of features for that class in the content images.

Building on the feature fusion method, we design the learning strategy of SSID: randomly sample and fuse features from different domains to generate labeled, style-rich features. This strategy enriches the feature space of each class and implicitly transfers class-discriminative information from the source to the target domain.

As shown in Fig.~\ref{fig_2}, SSID learning strategy is appended to each feature extractor block. The input feature maps of the \(l\)-th learning strategy in source domain, SSID and target domain denoted as \(f_s^l\), \(f_i^l\), and \(f_t^l \in {\mathbb{R}^{B \times N \times {C^l}}}\) respectively, where \(B\) represents the batch size, \(N\) represents the window size and \({C^l}\) is the output feature dimension in this block. 
In the SSID learning strategy, we first randomly sample the feature maps of the three domains at a one-third scale to obtain feature maps \(f_{sample}^l \in {\mathbb{R}^{B \times N \times {C^l}}}\). Then calculate its channel-wise mean \(\mu ^l\in {\mathbb{R}^{B \times {C^l}}}\) and variance \(\sigma ^l\in {\mathbb{R}^{B \times {C^l}}}\) as style features. These are followed by a \(1 \times 1\) convolution layer \(Conv\). Next, we multiply and add \(\mu^l\) and \(\sigma^l\) channel-wise by \(f_i^l\) as: 
 
\begin{equation}
\label{eq2}
f_i^l \leftarrow (f_i^l \bigotimes FC(\sigma ^l)) \bigoplus FC(\mu ^l),
\end{equation} 
where \(\bigotimes\) and \(\bigoplus\) represent channel multiplication and addition, respectively. By discarding the object features and utilizing only the style features of \(f_{sample}^l\), we preserve the semantic information and labels of \(f_i^l\). 
% Through continuous feature fusion, SSID can build a labeled, style-rich latent feature space for each class, so as to implicitly transfer class-discriminative information from the source to the target domain. 
Through continuous feature fusion, SSID constructs a labeled and style-rich latent feature space for each category. In this space, category labels are sourced from the source domain, while style information is derived from the target domain. This approach establishes an intermediate feature space that closely aligns with the target domain’s feature distribution, naturally inheriting labels without requiring manual annotation. The features in this intermediate domain strongly resemble those of the target domain. By training on the intermediate domain, SSID effectively transfers category-distinguishing information to the target domain, enhancing its recognition performance.

\subsection{Memory Bank}
\label{subsec32}

In our method, we utilize an external memory bank with six memory cells to store the specified features of the SSID and the source domain in real time during training via index. These features are then applied to calculate stable class features and class-wise style features for downstream computing tasks, as shown in Fig.~\ref{fig_2}. In this section, we detailed the workflow of our proposed memory bank.

\subsubsection{Initialization Strategy of Memory Bank}
Before training, the pre-trained feature extractor with frozen model parameters is utilized to initialize the memory bank as follows:

\begin{itemize}
\item{Since the output feature maps of the feature extractor \(F^L=\left[f_s^L,f_i^L,f_t^L\right]\) contains rich semantic information with style clues, and feature vector \(F^{cls}=\left[f_s^{cls},f_i^{cls},f_t^{cls} \right]\) before the classifier head layer contains refined class-discriminative information, we store \(f_s^L\in {\mathbb{R}^{B \times N \times {C^L}}}\) and \(f_i^{cls}\in {\mathbb{R}^{B \times {C^{cls}}}}\) in the memory bank, where \(C^L\) and \(C^{cls}\) denote the output feature dimension of feature extractor and bottleneck layer.}
\item{Calculate the channel-wise mean $\mu \left({f_s^L} \right) \in {\mathbb{R}^{B \times {C^L}}}$ and variance $\sigma \left({f_s^L} \right) \in {\mathbb{R}^{B \times {C^L}}}$  of \(f_s^L\), and update the corresponding \(B\) features in memory cells $M_\mu ^{{N_s}} \in {\mathbb{R}^{{N_s} \times {C^L}}}$ and $M_\sigma ^{{N_s}} \in {\mathbb{R}^{{N_s} \times {C^L}}}$ according to the sample index. Meanwhile, $M_{cls}^{{N_s}} \in {\mathbb{R}^{{N_s} \times {C^{cls}}}}$ is updated with \(f_i^{cls}\) according to the sample index.}
\item{Repeat the above steps until all samples in \(S\) and \(I\) are traversed. Calculate the class centers of all features in $\left\{ {M_\mu ^{{N_s}},M_\sigma ^{{N_s}},M_{cls}^{{N_s}}}\right\}$ according to their class labels and store the result in $\left\{ {M_\mu ^K \in {\mathbb{R}^{K \times {C^L}}}, M_\sigma ^K \in {\mathbb{R}^{K \times {C^L}}}, M_{cls}^K \in {\mathbb{R}^{K \times {C^{cls}}}}} \right\}$}, respectively. Finally, unfreeze the model parameters.
\end{itemize}

\subsubsection{Update Strategy of Memory Bank}

During training, the parameters of the feature extractor are continuously updated, and the most recent features are used to update the memory block. The process is as follows:
\begin{itemize}
\item{During forward propagation, the latest \(f_s^L\) and \(f_i^{cls}\) for the current epoch are extracted.}
\item{Calculate the channel-wise mean $\mu \left({f_s^L} \right) $ and variance $\sigma \left({f_s^L} \right)$, and the corresponding old values in $M_\mu ^{{N_s}}$ and $M_\sigma ^{{N_s}}$ are replaced based on the sample index. Simultaneously, \(f_i^{cls}\) is applied to update the corresponding value in $M_{cls}^{{N_s}}$.}
\item{Calculate the class centers of all features in $\left\{ {M_\mu ^{{N_s}},M_\sigma ^{{N_s}},M_{cls}^{{N_s}}} \right\}$ according to their class labels and store the result in $\left\{ {M_\mu ^K, M_\sigma ^K, M_{cls}^K} \right\}$, respectively.}
\end{itemize}

Furthermore, the proposed memory bank is only an external repository and does not participate in the back-propagation calculation of the network.

\subsection{Loss Function}
\label{subsec33}

Based on the designed memory bank, we propose a novel joint loss function that integrates the intra-domain loss ${\mathcal{L}_{intra}}$ and inter-domain loss ${\mathcal{L}_{inter}}$ to improve the class recognition ability and feature compatibility, respectively. The overall loss function $\mathcal{L}$ is as follows:

\begin{equation}
\label{eq3}
\mathcal{L} = {\mathcal{L}_{intra}} + \alpha \cdot {\mathcal{L}_{inter}},
\end{equation}
where \(\alpha\) is a hyperparameter with a default value of 1.0 to balance \({\mathcal{L}_{intra}}\) and \({\mathcal{L}_{inter}}\).

\subsubsection{Intra-domain Loss Function}

Intra-domain loss function \({\mathcal{L}_{intra}}\) is designed to improve the class recognition ability, which consists of the respective domain loss functions of the source domain \({\mathcal{L}_s}\), style-aware self-intermediate domain \({\mathcal{L}_i}\) and target domain \({\mathcal{L}_t}\), as:
\begin{equation}
\label{eq4}
{\mathcal{L}_{intra}} = {\mathcal{L}_s} + {\mathcal{L}_i} + {\mathcal{L}_t}.
\end{equation}

For the labeled source domain and SSID, the network can be trained with the traditional cross-entropy loss function:
\begin{align}
\label{eq5}
{\mathcal{L}_s} = {\mathcal{L}_{CE}}\left( {y_s^{pre},{y_s}} \right), \\
\label{eq6}
{\mathcal{L}_i} = {\mathcal{L}_{CE}}\left( {y_i^{pre},{y_i}} \right),
\end{align} 
where \(\mathcal{L}_{CE}\) represent the cross-entropy loss function, \(y^{pre}\) and \(y\) denote the predicted result by classifier head layer and the ground truth, respectively. Meanwhile, mutual information maximization is employed on the unlabeled target domain, as: 
\begin{equation}
\label{eq7}
{\mathcal{L}_t} = \frac{1}{{{N_t}}}\sum\limits_{p = 1}^{{N_t}} {\sum\limits_{k = 1}^K {y_{tpc}^{pre}} } \log (y_{tpc}^{pre}) - \sum\limits_{k = 1}^K {y_{tc}^{pre}\log (y_{tc}^{pre})},
\end{equation} 
where \(y_{tpc}^{pre}\) denotes the  probability that the \(p\)-th sample in the target domain is predicted to be the \(c\)-th class, and \(y_{tc}^{pre} = \sum\limits_{p = 1}^{{N_t}} {y_{tc}^{pre}}\).

\subsubsection{Inter-domain Loss Function}

Inter-domain loss function \({\mathcal{L}_{inter}}\) consists of discrimination loss function \({\mathcal{L}_D}\) and style loss function \({\mathcal{L}_{style}}\), as:
\begin{equation}
\label{eq8}
{\mathcal{L}_{inter}} = {\mathcal{L}_D} + {\mathcal{L}_{style}}.
\end{equation}

To maintain the consistency of the class features of the source domain $\&$ SSID and the target domain $\&$ SSID to improve the discriminability, we design a discrimination loss function \(\mathcal{L}_D\), as:
\begin{equation}
\label{eq9}
{\mathcal{L}_D} = KL(M_{cls}^K[y_s^{pre}]||f_s^{cls}) + KL(M_{cls}^K[y_t^{pre}]||f_t^{cls}),
\end{equation} 
where \(f_s^{cls}\) and \(f_t^{cls}\) represent the real-time feature vectors of the source domain and the target domain before the classifier head layer in the current iteration, respectively. \(M_{cls}^K[y^{pre}]\) represents the class center of \(y^{pre}\) stored in memory cell \(M_{cls}^K\), and \(KL\) stands for Kullback–Leibler divergence.  

Style loss function  \({\mathcal{L}_{style}}\) aims to reduce the difference of class-wise style features in the source and target domain to improve inter-domain transferability, as:
\begin{equation}
\label{eq10}
\begin{aligned}
& {\mathcal{L}_{style}} = {up_{rate}} \cdot KL(M_\mu ^K[y_t^{pre}]||\mu (f_t^L))  \\
& \quad \quad \quad + KL(M_\sigma ^K[y_t^{pre}]||\mu (f_t^L)),
\end{aligned}
\end{equation}  
where \(M_\mu ^K[y_t^{pre}]\) and \(M_\sigma ^K[y_t^{pre}]\) represents the style features of \(y_t^{pre}\) stored in memory cell \(M_\mu^K\) and \(M_\sigma^K\) respectively. \(f_t^L\) denotes the feature maps of the target domain output by the feature extractor in the current iteration, \(\mu ( \cdot )\) and \(\sigma ( \cdot )\) denote mean and variance operations, respectively. Furthermore, since the style features in the early stage of the network are not robust enough, we introduce a rising factor \({up_{rate}} = {\left( {\frac{{epoch\_num}}{{total\_epoch}}} \right)^{rate}}\), where \(rate\) is set to 0.9. This choice strikes a balance between stability, smooth growth, and sensitivity, allowing the rising factor to grow more gradually in the later stages of training and preventing instability from prematurely focusing on style matching.
 
\subsubsection{Convergence Analysis of Loss Function}
The convergence of the loss function is a crucial necessary condition for calculating the optimal network parameters. With a limited number of samples, it is easy to infer that the cross-entropy loss function and the mutual information maximization function are convergent, that is, \(\mathcal{L}_s\), \(\mathcal{L}_t\), \(\mathcal{L}_D\) and \(\mathcal{L}_{style}\) are all convergent. Therefore, in this section, we focus on proving the convergence of \(\mathcal{L}_i\). The expansion of \({\cal L}_i\) is as follows:
\begin{equation}
\label{eq13_1}
\mathcal{L}_i = \frac{1}{N_s}\sum\limits_{n = 1}^{N_s}\frac{1}{N_i}\sum\limits_{m = 1}^{N_i} { - y_{in} \log (y_{in \rightarrow m}^{pre})}.
\end{equation} 
where \(N_i\) denotes the number of features in the latent feature space of SSID. \(y_{in}\) and \(y_{in \rightarrow m}^{pre}\) represent the ground truth of the \(n\)-th input sample and the prediction result of the \(m\)-th feature fusion of the n-th sample, respectively.

From a statistical point of view, the latent feature space of SSID is huge. At each iteration, the mini-batch of each domain is randomly selected, a total \({\left( {C_{{N_s}}^B} \right)^2}C_{{N_t}}^B{  ({A_B^{B/3}})^3 }\) features can be obtained after one layer of SSID learning strategy. After \(L\) layers, the number of features \({N_i}\) in the latent feature space will reach a large value. Since it is nearly impossible to traverse all features in the latent feature space of the SSID, the real number \(N_i\) is approximately replaced by the sampling number \(M\). 

Assuming that the feature of each class in each domain obeys a normal distribution \(N(\mu _d^k,\sigma _d^k)\) , where \(k\) denote the class number, and \(d\) takes the value from \(\left[s,i,t \right]\). For the input sample \((x_{in},y_{in} )\) in SSID, its fused feature before the classifier head layer is noted as \(\widehat{f_{in \rightarrow 1}^{{y_{in}}}}\). After \(M\) times sampling and fusion, an auxiliary synthesis feature set of \((x_{in},y_{in} )\) will be obtained:
\begin{equation}
\label{eq11}
\left\{ {\left( {\widehat {f_{in\rightarrow 1}^{y_{in}}},y_{in}} \right),\left( {\widehat {f_{in\rightarrow 2}^{y_{in}}},y_{in}} \right), \ldots ,\left( {\widehat {f_{in\rightarrow M}^{y_{in}}},y_{in}} \right)} \right\}.
\end{equation} 
Under the above condition, using the parameters of the classifier head layer and softmax to expand the predicted value \(y_{in \rightarrow m}^{pre}\) of all input samples, then Eq.~(\ref{eq13_1}) can be rewritten as: 
\begin{equation}
\label{eq12}
\mathcal{L}_i \left( {\theta ,W,b} \right) = \frac{1}{N_s}\sum\limits_{n = 1}^{N_s}\frac{1}{M}\sum\limits_{m = 1}^M { - \log (\frac{{{e^{W_{{y_{in}}}^T \widehat {f_{in\rightarrow m}^{{y_{in}}}} + {b_{{y_{in}}}}}}}}{{\sum\limits_{k' = 1}^K {{e^{W_{k'}^T\widehat {f_{in\rightarrow m}^{{y_{in}}}} + {b_{k'}}}}} }})},
\end{equation} 
where \(W\) and \(b\) denote the weight matrix and bias vector of the classifier head layer respectively. Eq.~(\ref{eq12}) aims to reduce the difference between the predicted distribution and the actual distribution of features in the latent feature space of the SSID, so as to improve the classification performance of the network. Intuitively, the closer the value of \(M\) is to \(N_i\), the more robust the features in the latent space. Therefore, we need to explore whether \(\mathcal{L}_i\) converges when the sampling times \(M\) are continuously growing until infinity.
When \(M\rightarrow\infty\), Eq.~(\ref{eq12}) can be rewritten as: 
\begin{equation}
\label{eq13}
\begin{aligned}
& \mathop {\lim }\limits_{M \to \infty } {\mathcal{L}_i}\left( {\theta ,W,b} \right)  \\
& = \mathop {\lim }\limits_{M \to \infty } \frac{1}{N_s}\sum\limits_{n = 1}^{N_s}\frac{1}{M}\sum\limits_{m = 1}^M { - \log (\frac{{{e^{W_{{y_{in}}}^T \widehat {f_{in\rightarrow m}^{{y_{in}}}} + {b_{{y_{in}}}}}}}}{{\sum\limits_{k' = 1}^K {{e^{W_{k'}^T\widehat {f_{in\rightarrow m}^{{y_{in}}}} + {b_{k'}}}}} }})} \\
& = \frac{1}{{{N_s}}}\sum\limits_{n = 1}^{{N_s}} {{E_{\widehat {f_{in \to m}^{{y_{in}}}}}}} \left[ { - \log (\frac{{{e^{W_{{y_{in}}}^T\widehat {f_{in \to m}^{{y_{in}}}} + {b_{{y_{in}}}}}}}}{{\sum\limits_{k' = 1}^K {{e^{W_{k'}^T\widehat {f_{in \to m}^{{y_{in}}}} + {b_{k'}}}}} }})} \right] \\
& = \frac{1}{{{N_s}}}\sum\limits_{n = 1}^{{N_s}} {\sum\limits_{k = 1}^K {\frac{{{N_{ik}}}}{{{N_i}}}} {E_{\widehat {f_{in}^k}}}[\log \left( {\sum\limits_{k' = 1}^K {{e^{\left( {W_{k'}^T - W_{{y_{in}}}^T} \right)\widehat {f_{in}^k} + {b_{k'}} - {b_{{y_{in}}}}}}} } \right)]},
\end{aligned}
\end{equation} 
where \(\frac{N_{ik}}{N_i}\) denotes the proportion of the \(k\)-th class in the latent feature space, which is unknown but can be estimated statistically. According to the moment-generating function and Jason's inequality, the upper bound of Eq.~(\ref{eq13}) can be found as follows:
\begin{equation}
\label{eq14}
\begin{aligned}
& \mathop {\lim }\limits_{M \to \infty } {\mathcal{L}_i }\left( {\theta ,W,b} \right) \\
& = \frac{1}{{{N_s}}}\sum\limits_{n = 1}^{{N_s}} {\sum\limits_{k = 1}^K {\frac{{{N_{ik}}}}{{{N_i}}}} {E_{\widehat {f_{in}^k}}}[\log \left( {\sum\limits_{k' = 1}^K {{e^{\left( {W_{k'}^T - W_{{y_{in}}}^T} \right)\widehat {f_{in}^k} + {b_{k'}} - {b_{{y_{in}}}}}}} } \right)]} \\
& = \frac{1}{{{N_s}}}\sum\limits_{n = 1}^{{N_s}} {\sum\limits_{k = 1}^K {\frac{{{N_{ik}}}}{{{N_i}}}} {E_{\widehat {f_{in}^k}}}[\log \left( {\sum\limits_{k' = 1}^K {{e^{\left( {W_{k'y}} \right)\widehat {f_{in}^k} + B}}}} \right)]} \\
&  \le \frac{1}{{{N_s}}}\sum\limits_{n = 1}^{{N_s}} {\sum\limits_{k = 1}^K {\frac{{{N_{ik}}}}{{{N_i}}}} \log \left( {\sum\limits_{k' = 1}^K {{E_{\widehat {f_{in}^k}}}} \left[ {{e^{W_{k'y} \widehat {f_{in}^k} + B}}} \right]} \right)} \\
& = \frac{1}{{{N_s}}}\sum\limits_{n = 1}^{{N_s}} {\sum\limits_{k = 1}^K {\frac{{{N_{ik}}}}{{{N_i}}}} \log \left( {\sum\limits_{k' = 1}^K {{e^{W_{k'y}\widehat {\mu _{in}^k} + B + \frac{1}{2}{W_{k'y}^T}\widehat {\sigma _{in}^k}W_{k'y}}}} } \right)} \\
& = \frac{1}{{{N_s}}}\sum\limits_{n = 1}^{{N_s}} {\sum\limits_{k = 1}^K {\frac{{{N_{ik}}}}{{{N_i}}}} \left( { - \log \left( {\frac{{{e^{{A_{{y_{in}}}}}}}}{{\sum\limits_{k' = 1}^K {{e^{{A_{k'y}}}}} }}} \right)} \right)},
\end{aligned}
\end{equation} 
where \(W_{k'y} = W_{k'}^T - W_{{y_{in}}}^T\) and \(B = b_{k'} - b_{{y_{in}}}\), respectively. \({A_{{y_{in}}}} = W_{{y_{in}}}^T\widehat {\mu _{in}^k} + {b_{{y_{_{in}}}}} + \frac{1}{2}{(W_{{y_{in}}}^T - W_{{y_{in}}}^T)^T}\widehat {\sigma _{in}^k}(W_{{y_{in}}}^T - W_{{y_{in}}}^T)\) and \({A_{k'}} = W_{k'}^T\widehat {\mu _{in}^k} + {b_{k'}} + \frac{1}{2}{(W_{k'}^T - W_{{y_{in}}}^T)^T}\widehat {\sigma _{in}^k}(W_{k'}^T - W_{{y_{in}}}^T)\) are polynomials related to \(W_{y_{in}}^T\) and \(W_{k'}^T\). 

Essentially, since the upper bound function Eq.~(\ref{eq14}) is a form of cross-entropy loss function, \(\mathop {\lim }\limits_{M \to \infty } {L_i}\left( {\theta ,W,b} \right)\) is convergent and its performance is independent of the value of the sampling times \(M\). Finally, we summarize the strategy of our proposed SSID in~\ref{app1}.

\begin{table}
\begin{center}
  \caption{Accuracy ($\%$) of Comparative Experiments on VisDA-2017. The Bold Numbers Indicate the Best Performance.}
  \label{tab:Table1}
  \resizebox{0.6\textwidth}{!}{
  \begin{tabular}{cc|cc}
    \toprule
    Methods & Sy \(\rightarrow\) Re & Methods & Sy \(\rightarrow\) Re \\
    \midrule

    DAN~\cite{ref6}        & 61.1 & MODEL~\cite{ref74}      & 81.6 \\
    BNM~\cite{ref15}       & 70.4 & CRAT+CaCo~\cite{ref71}  & 81.6 \\
    MCD~\cite{ref19}       & 71.9 & BSP+TSA~\cite{ref34}    & 82.0 \\
    ViT-B~\cite{ref38}     & 72.6 & STAR~\cite{ref73}       & 82.7 \\
    DeiT-B~\cite{ref57}    & 73.2 & SHOT~\cite{ref56}       & 82.9 \\
    CDAN+E~\cite{ref17}    & 73.9 & TransDA~\cite{ref36}    & 83.0 \\
    RWSAN~\cite{ref69}     & 73.9 & MCC+NWD~\cite{ref62}    & 83.7 \\
    CaCo~\cite{ref71}      & 80.9 & SDAT~\cite{ref76}       & 84.3 \\
    TransConv~\cite{ref70} & 80.9 & MSGD~\cite{ref75}       & 84.6 \\
    PACMAC~\cite{ref72}    & 81.0 & TVT + SSID            & \bf 85.4 \\
  \bottomrule
\end{tabular}}
\end{center}
\end{table}

\section{Experiment}
\subsection{Implementation Details}
We evaluate the proposed method on two challenging UDA benchmarks. One is VisDA-2017~\cite{ref45}, a Synthetic-to-Real dataset comprising 152,397 synthetic domain (Sy) images and 55,388 real domain (Re) images across 12 categories. The second is Office-Home~\cite{ref46}, which includes 15,500 images from 65 categories in four domains: Artistic (A), Clipart (C), Product (P), and RealWorld (R) images. 
% Following the standard protocol, we employ accuracy as the performance metric for each task.
For a fair comparison, We calculate accuracy following the approach described in~\cite{ref17,ref34}: 1) For the VisDA-2017~\cite{ref45}, the per-class accuracy $A_k=\frac{correct\_prediction\_of\_class\_k}{total\_sample\_of\_class\_k}$ is computed, and the overall average accuracy is given by $A=\frac{1}{K} \sum\limits_{k = 1}^{K} A_k$. 2) For the Office-Home~\cite{ref46}, the total accuracy is calculated as: $A=\frac{correct\_prediction}{total\_sample}$.

We utilize several mainstream vision transformers, such as SWIN~\cite{ref47}, BOAT~\cite{ref48}, and TVT~\cite{ref37}, with a patch size of 16×16, as the feature extractors to assess the effectiveness of the proposed SSID. 
% The number of feature extractor blocks \(L\) is set to 4, matching the number of stages in both SWIN and BOAT. For TVT, which has twelve stages, we group its stages into four blocks, aligning with SSID's structure. 
We divide the structure of the feature extractor based on the feature dimensions, with SWIN, BOAT, and TVT each consisting of four feature extraction blocks, i.e., \(L=4\).
% Pre-trained models were used for all three backbones on all datasets to ensure a fair comparison. 
To ensure a fair comparison, all three backbones use pre-trained models: “SWIN-T” for SWIN and BOAT, and “ViT-B” for TVT, across all datasets.
All experiments are conducted using PyTorch on an NVIDIA Tesla K40 GPU with 12GB memory, with a batch size of 18 for each domain.

\begin{table}
\begin{center}

  \caption{Accuracy ($\%$) of Backbone Experiments on VisDA-2017. The Bold Numbers Indicate the Best Performance. Statistical significance (p-value $<$ 0.05~\cite{p2}) is denoted with: $^{\diamond}$.}
  \label{tab:Table2}
  \resizebox{0.6\textwidth}{!}{
  \begin{tabular}{cc|cc}
    \toprule
    Methods & Sy \(\rightarrow\) Re & Methods & Sy \(\rightarrow\) Re \\
    \midrule
    SWIN (backbone) & 74.2 & SWIN + SSID & \bf 81.7$^{\diamond}$\\    
    BOAT (backbone) & 78.6 & BOAT + SSID & \bf 84.0$^{\diamond}$\\    
    TVT (backbone)  & 84.8 &  TVT + SSID & \bf 85.4$^{\diamond}$\\
  \bottomrule
\end{tabular}}
\end{center}

\end{table}

\begin{table*}
  \caption{Accuracy ($\%$) of Comparative Experiments on Office-Home. The Bold Numbers Indicate the Best Performance.}
  \label{tab:Table3}
    \resizebox{0.99\textwidth}{!}{
  \begin{tabular}{c|cccccccccccc|c}
    \toprule
    \multirow{3}{*}{Method} & \multirow{3}{*}{\begin{tabular}[c]{@{}c@{}} A \\ \(\downarrow\) \\ C \end{tabular}} & \multirow{3}{*}{\begin{tabular}[c]{@{}c@{}} A \\ \(\downarrow\) \\P \end{tabular}} & \multirow{3}{*}{\begin{tabular}[c]{@{}c@{}} A \\ \(\downarrow\) \\ R \end{tabular}} & \multirow{3}{*}{\begin{tabular}[c]{@{}c@{}} C \\ \(\downarrow\) \\ A \end{tabular}} & \multirow{3}{*}{\begin{tabular}[c]{@{}c@{}} C \\ \(\downarrow\) \\P \end{tabular}} & \multirow{3}{*}{\begin{tabular}[c]{@{}c@{}} C \\ \(\downarrow\) \\ R \end{tabular}} & \multirow{3}{*}{\begin{tabular}[c]{@{}c@{}} P \\ \(\downarrow\) \\ A \end{tabular}} & \multirow{3}{*}{\begin{tabular}[c]{@{}c@{}} P \\ \(\downarrow\) \\ C \end{tabular}} & \multirow{3}{*}{\begin{tabular}[c]{@{}c@{}} P \\ \(\downarrow\) \\ R \end{tabular}} & \multirow{3}{*}{\begin{tabular}[c]{@{}c@{}} R \\ \(\downarrow\) \\ A \end{tabular}} & \multirow{3}{*}{\begin{tabular}[c]{@{}c@{}} R \\ \(\downarrow\) \\ C \end{tabular}} & \multirow{3}{*}{\begin{tabular}[c]{@{}c@{}} R \\ \(\downarrow\) \\ P \end{tabular}} &  \multirow{3}{*}{Avg} \\
    & \multicolumn{12}{c|}{} \\
    & \multicolumn{12}{c|}{} \\
    \midrule
    CDAN+E~\cite{ref17}      & 50.7 & 70.6 & 76.0 & 57.6 & 70.0 & 70.0 & 57.4 & 50.9 & 77.3 & 70.9 & 56.7 & 81.6 & 65.8\\
    GVB-GD~\cite{ref67}      & 57.0 & 74.7 & 79.8 & 64.6 & 74.1 & 74.6 & 65.2 & 55.1 & 81.0 & 74.6 & 59.7 & 84.3 & 70.4\\
    BSP+TSA~\cite{ref34}     & 57.6 & 75.8 & 80.7 & 64.3 & 76.3 & 75.1 & 66.7 & 55.7 & 81.2 & 75.7 & 61.9 & 83.8 & 71.2\\
    SHOT~\cite{ref56}        & 57.1 & 78.1 & 81.5 & 68.0 & 78.2 & 78.1 & 67.4 & 54.9 & 82.2 & 73.3 & 58.8 & 84.3 & 71.8\\
    GVB-GD+SPCL~\cite{ref60} & 59.3 & 76.8 & 80.7 & 66.6 & 75.9 & 75.8 & 66.7 & 57.4 & 82.9 & 76.5 & 61.8 & 85.8 & 72.2\\
    MCC+NWD~\cite{ref62}     & 58.1 & 79.6 & 83.7 & 67.7 & 77.9 & 78.7 & 66.8 & 56.0 & 81.9 & 73.9 & 60.9 & 86.1 & 72.6\\
    Deit-B~\cite{ref57}      & 61.8 & 79.5 & 84.3 & 75.4 & 78.8 & 81.2 & 72.8 & 55.7 & 84.4 & 78.3 & 59.3 & 86.0 & 74.8 \\
    RWSAN~\cite{ref69}       & 63.1 & 81.6 & 84.1 & 71.4 & 69.7 & 81.9 & 71.8 & 59.2 & 88.8 & 80.6 & 60.9 & 88.8 & 75.1 \\
    MCC+GLBA-S~\cite{ref77}  & 66.7 & 82.8 & 86.2 & 76.9 & 81.5 & 82.3 & 76.6 & 61.8 & 86.0 & 80.8 & 65.8 & 88.9 & 78.0\\
    TransDA~\cite{ref36}     & 67.5 & 83.3 & 85.9 & 74.0 & 83.8 & 84.4 & 77.0 & 68.0 & 87.0 & 80.5 & 69.9 & 90.0 & 79.3\\
    ViT-B~\cite{ref38}       & 67.0 & 85.7 & 88.1 & 80.1 & 84.1 & 86.7 & 79.5 & 67.0 & 89.4 & 83.6 & 70.2 & 91.2 & 81.1\\
    VP~\cite{ref78}          & 66.7 & 89.1 & 89.1 & 81.7 & 89.0 & 89.2 & 81.8 & 67.0 & 89.1 & 81.7 & 66.6 & 89.0 & 81.7\\
    TransConv~\cite{ref70}   & 69.9 & 87.1 & 88.6 & 82.6 & 87.5 & 88.4 & 82.1 & 70.2 & 89.8 & 84.6 & 73.1 & 91.1 & 82.9\\
    DOT-B~\cite{ref3700}     & 73.1 & 89.1 & 90.1 & \bf 85.5 & \bf 89.4 & \bf 89.6 & 83.2 & 72.1 & 90.4 & 84.4 & 72.9 & \bf 91.5 & 84.3 \\
    TVT + SSID              & \bf 77.0 & \bf 89.3 & \bf 90.3 & 85.3 & \bf 89.4 & 89.4 & \bf 83.8 & \bf 73.8 & \bf 91.1 & \bf 86.3 & \bf 75.6 & \bf 91.5 & \bf 85.3\\
  \bottomrule
\end{tabular}}
\end{table*}

\subsubsection{Results on VisDA-2017 Dataset}
VisDA-2017 is a challenging dataset with large domain discrepancies for UDA. We compare the proposed method with 15 SOTA methods on this dataset, including adversarial-based, difference-based, and mixed methods, as reported in Tables~\ref{tab:Table1} and~\ref{tab:Table2}. The left side of "\(\rightarrow\)" represents the source domain (training set), while the right side represents the target domain (testing set).

As shown in Table~\ref{tab:Table1}, TVT+SSID achieves the best performance, proving the effectiveness of the proposed SSID. DAN~\cite{ref6} achieves the worst results, likely due to its use of simple specific layers to learn invariant features between domains, which may weaken category differences and result in lower recognition ability. MCD~\cite{ref19}, CDAN+E~\cite{ref17} and SDAT~\cite{ref76} further exploit discriminative information conveyed in the classifier predictions to assist adversarial adaptation, improving classification accuracy by 17.7\%, 20.9\% and 38.9\% respectively. 

Compared with the adversarial-based methods, the difference-based methods aim to align features between domains by minimizing the difference of prototypes, with TransDA~\cite{ref36} achieving an accuracy rate of 83.0\%. MSGD~\cite{ref75} ranks second with 84.6\% but requires complex structural knowledge construction after classifier. In contrast, SSID introduces a novel approach that synthesizes object-invariant features with labels by utilizing information from all domains. This leads to better alignment of object regions, improved inter-domain generalization, and enhanced robustness. TVT+SSID achieves an accuracy rate of 85.40\%, outperforming other methods.
BSP+TSA~\cite{ref34} achieves an accuracy of 82.0\% by enriching internal feature patterns in the latent space through random sampling enhancement. However, However, unlike SSID, its enhancement approach is unidirectional and does not consider enhancing information between different classes.

Additionally, SSID consistently improves the performance of three mainstream vision transformer-based backbones, with SWIN, BOAT, and TVT gaining 10.22\%, 6.90\%, and 0.99\% improvement, respectively, as demonstrated in Table~\ref{tab:Table2}. These results indicate that SSID effectively enhances the feature transferability and discriminability of the classifier on this challenging cross-domain dataset.

\begin{table*}
  \caption{Accuracy ($\%$) of Backbone Experiments on Office-Home. The Bold Numbers Indicate the Best Performance. Statistical significance (p-value $<$ 0.05~\cite{p2}) is denoted with: $^{\diamond}$.}
  \label{tab:Table4}
  \resizebox{0.99\textwidth}{!}{
  \begin{tabular}{c|cccccccccccc|c}
    \toprule
    \multirow{3}{*}{Method} & \multirow{3}{*}{\begin{tabular}[c]{@{}c@{}} A \\ \(\downarrow\) \\ C \end{tabular}} & \multirow{3}{*}{\begin{tabular}[c]{@{}c@{}} A \\ \(\downarrow\) \\P \end{tabular}} & \multirow{3}{*}{\begin{tabular}[c]{@{}c@{}} A \\ \(\downarrow\) \\ R \end{tabular}} & \multirow{3}{*}{\begin{tabular}[c]{@{}c@{}} C \\ \(\downarrow\) \\ A \end{tabular}} & \multirow{3}{*}{\begin{tabular}[c]{@{}c@{}} C \\ \(\downarrow\) \\P \end{tabular}} & \multirow{3}{*}{\begin{tabular}[c]{@{}c@{}} C \\ \(\downarrow\) \\ R \end{tabular}} & \multirow{3}{*}{\begin{tabular}[c]{@{}c@{}} P \\ \(\downarrow\) \\ A \end{tabular}} & \multirow{3}{*}{\begin{tabular}[c]{@{}c@{}} P \\ \(\downarrow\) \\ C \end{tabular}} & \multirow{3}{*}{\begin{tabular}[c]{@{}c@{}} P \\ \(\downarrow\) \\ R \end{tabular}} & \multirow{3}{*}{\begin{tabular}[c]{@{}c@{}} R \\ \(\downarrow\) \\ A \end{tabular}} & \multirow{3}{*}{\begin{tabular}[c]{@{}c@{}} R \\ \(\downarrow\) \\ C \end{tabular}} & \multirow{3}{*}{\begin{tabular}[c]{@{}c@{}} R \\ \(\downarrow\) \\ P \end{tabular}} &  \multirow{3}{*}{Avg} \\
    & \multicolumn{12}{c|}{} \\
    & \multicolumn{12}{c|}{} \\
    \midrule
    SWIN   (backbone) & 58.8 & 77.7 & 81.4 & 69.8 & 76.3 & 78.3 & 68.2 & 55.1 & 82.5 & 74.6 & 60.9 & 84.3 & 72.3 \\
    \bf SWIN + SSID   & \bf 58.9 & \bf 78.6 & \bf 82.2 & \bf 70.8 & \bf 78.2 & \bf 79.2 & \bf 69.1 & \bf 55.1 & \bf 83.0 & \bf 75.9 & \bf 61.6 & \bf 85.2 & \bf 73.2$^{\diamond}$ \\    \hline
    BOAT   (backbone) & 62.3 & 81.2 & 84.6 & 76.1 & 80.2 & 81.4 & 74.7 & 59.1 & 83.9 & 78.5 & 64.1 & 87.4 & 76.1 \\
    \bf BOAT+ SSID    & \bf 63.9 & \bf 81.3 & \bf 85.2 & \bf 76.5 & \bf 81.4 & \bf 81.4 & \bf 75.1 & \bf 59.7 & \bf 85.0 & \bf 80.0 & \bf 64.4 & \bf 87.7 & \bf 76.8$^{\diamond}$ \\    \hline
    TVT   (backbone)  & 75.0 & 87.0 & 89.5 & 82.5 & 87.8 & 88.2 & 82.2 & 73.5 & 90.1 & 85.0 & 74.8 & 90.9 & 83.9 \\
    \bf TVT + SSID    & \bf 77.0 & \bf 89.3 & \bf 90.3 & \bf 85.3 & \bf 89.4 & \bf 89.4 & \bf 83.8 & \bf 73.8 & \bf 91.1 & \bf 86.3 & \bf 75.6 & \bf 91.5 & \bf 85.3$^{\diamond}$\\
  \bottomrule
\end{tabular}}
\end{table*}

\subsubsection{Results on Office-Home Dataset}
To further assess its generality, we validate the proposed method on the Office-Home dataset and compare its performance with 14 SOTA methods, as shown in Table~\ref{tab:Table3}. 
% The Office-Home dataset consists of four domains forming 12 different UDA tasks.
The Office-Home dataset consists of four domains. We select one domain to serve as the training set (The left side of "\(\rightarrow\)") and another domain as the test set (The right side of "\(\rightarrow\)"), creating a total of 12 distinct UDA tasks. 
TVT+SSID achieves the best performance on all tasks, with an average accuracy of 85.3\%, significantly outperforming other UDA methods.

Furthermore, Table~\ref{tab:Table4} summarizes that SSID significantly enhances the performance of different backbones on all tasks. The average accuracies of SWIN+SSID, BOAT+SSID, and TVT+SSID are 73.16\%, 76.79\%, and 84.86\%, respectively. These results demonstrate that the proposed SSID can generate more transferable cross-domain representations and implicitly preserve class-discriminative information.

\begin{table}
\begin{center}

  \caption{Accuracy ($\%$) of Ablation Experiments on VisDA-2017. The Bold Numbers Indicate the Best Performance.}
  \label{tab:Table5}
  \resizebox{0.5\textwidth}{!}{
  \begin{tabular}{c|cccc|c}
    \toprule
    Methods &  \(\mathcal{L}_s+\mathcal{L}_t\) & \(\mathcal{L}_i\) & \(\mathcal{L}_D\) & \(\mathcal{L}_{style}\) & Sy \(\rightarrow\) Re\\
    \midrule
    \multirow{5}{*}{SWIN} & \checkmark & & & & 74.2\\
                          & \checkmark & \checkmark & & & 75.6\\
                          & \checkmark & \checkmark & \checkmark & & 78.2\\
                          & \checkmark & \checkmark & & \checkmark & 79.5\\
                          & \checkmark & \checkmark & \checkmark & \checkmark & \bf 81.7\\
    \cline{1-6}
    \multirow{5}{*}{BOAT} & \checkmark & & & & 78.6\\
                          & \checkmark & \checkmark & & & 80.8\\
                          & \checkmark & \checkmark & \checkmark & & 83.0\\
                          & \checkmark & \checkmark & & \checkmark & 81.0\\
                          & \checkmark & \checkmark & \checkmark & \checkmark & \bf 84.0\\
    \cline{1-6}
    \multirow{5}{*}{TVT}  &  \checkmark & & & & 84.8\\
                          & \checkmark & \checkmark & & & 85.1\\
                          & \checkmark & \checkmark & \checkmark & & 85.1\\
                          & \checkmark & \checkmark & & \checkmark & 85.2\\
                          & \checkmark & \checkmark & \checkmark & \checkmark & \bf 85.4\\
  \bottomrule
\end{tabular}}
\end{center}

\end{table}

\begin{table*}
  \caption{Accuracy ($\%$) of Experiments with Various \(\alpha\) on VisDA-2017. The Bold Numbers Indicate the Best Performance.}
  \label{tab:Table6}
  \resizebox{0.99\textwidth}{!}{
  \begin{tabular}{c|c|c|ccccccccc}
    \toprule
    Methods &  Backbone & Methods & \(\alpha=0.1\) & \(\alpha=0.3\) & \(\alpha=0.5\) & \(\alpha=0.7\) & \(\alpha=0.9\) & \(\alpha=1.0\)& \(\alpha=1.1\)& \(\alpha=1.3\)& \(\alpha=1.5\) \\
    \midrule
    SWIN   & 74.2     & SWIN + SSID & 76.3  & 77.6  & 77.5  & 79.0  & 81.3  & \bf 81.7 & 81.6 & 80.9 & 80.3 \\ 
    BOAT   & 78.6     & BOAT + SSID & 81.7  & 82.2  & 82.9  & 83.1  & 83.5  & \bf 84.0 & 83.0 & 83.0 & 82.9 \\ 
    TVT    & 84.8     & TVT + SSID  & 84.9  & 84.9  & 85.0  & 85.0  & 85.2  & \bf 85.4 & 85.2 & 84.8 & 84.8 \\ 
  \bottomrule
\end{tabular}}
\end{table*}

\begin{table*}
  \caption{Accuracy ($\%$) of Experiments with Various \(rate\) on VisDA-2017. The Bold Numbers Indicate the Best Performance.}
  \label{tab:Table7}
  \resizebox{0.99\textwidth}{!}{
  \begin{tabular}{c|cccccc}
    \toprule
    Methods & \(rate=0.5\) & \(rate=0.7\) & \(rate=0.9\) & \(rate=1.0\) & \(rate=1.1\) & \(rate=1.3\) \\
    \midrule
    SWIN + SSID & 77.3 & 78.9 & \bf 81.7 & 81.0 & 80.4 & 80.1 \\ 
    BOAT + SSID & 81.7 & 82.7 & \bf 84.0 & 83.5 & 82.9 & 82.7 \\ 
    TVT  + SSID & 84.9 & 83.9 & \bf 85.4 & 84.9 & 84.4 & 84.2 \\ 
  \bottomrule
\end{tabular}}
\end{table*}

\subsection{Analysis}
\subsubsection{Ablation Study}
As demonstrated in Table~\ref{tab:Table5}, this section investigates the contributions of each component in SSID and its associated loss functions. Specifically, \(\mathcal{L}_s+\mathcal{L}_t\) relates to the backbone, and \(\mathcal{L}_i\) is related to SSID. \(\mathcal{L}_D\) and \(\mathcal{L}_{style}\) are linked to memory bank, and are used to maintain class-discriminative information and improve style compatibility, respectively.

Table~\ref{tab:Table5} is divided into three sub-tables, corresponding to the adoption of SWIN, BOAT, and TVT backbones. The symbol '$\checkmark$' denotes the utilization of the loss function associated with the first row in Table~\ref{tab:Table5}. The first row in each sub-table yields the poorest results as it solely relies on the original backbone configuration.
Comparatively, the second row in each sub-table integrates SSID with the backbone, enriching the latent hypothesis space's features and increasing accuracy by 2.0\%, 2.8\%, and 0.4\% for SWIN, BOAT, and TVT, respectively. Additionally, integrating \(\mathcal{L}_D\) and \(\mathcal{L}_{style}\) related to the memory bank in the third and fourth rows of each sub-table further enhances the network's performance.
The final row in each sub-table represents the complete framework, surpassing any incomplete combination. This illustrates that each component of SSID can enhance various backbone models' performance, with these components synergistically boosting the model's overall performance. Furthermore, SSID is a lightweight method that imposes a negligible computational burden on the network.

\subsubsection{Hyperparameters}

We investigate the impact of the hyperparameter \(\alpha \) in the overall loss function $\mathcal{L}$ on network performance, as shown in Table~\ref{tab:Table6}. \(\alpha\) is the trade-off hyperparameter between the intra-domain loss function \({\mathcal{L}_{intra}}\) and the inter-domain loss function \({\mathcal{L}_{inter}}\), where \(\alpha<1\) indicates that \({\mathcal{L}_{intra}}\) is more important than \({\mathcal{L}_{inter}}\), and vice versa. 

As seen in Table~\ref{tab:Table6}, as \(\alpha \) approaches 1, the network performance improves, with the performance gain increasing continuously. Each backbone achieves its maximum performance at \(\alpha = 1.0\).  This indicates that the improvement in network performance is consistent and not dependent on a specific hyperparameter value. Therefore, in other experiments, we set \(\alpha \) to 1.0 by default, suggesting equal importance of intra- and inter-domain loss functions.

We further examine the influence of $rate$ on network performance, as presented in Table~\ref{tab:Table7}. As shown, the choice of rate directly affects the growth dynamics of the rising factor $up_rate$. During the early training stages, feature learning remains unstable, particularly in style feature extraction, which may lack precision. Assigning excessive weight to style loss at this stage (e.g., $rate=0.5$) could cause the model to prioritize style matching prematurely, at the expense of learning fundamental features. This may introduce instability during training and negatively impact the final task performance. Conversely, a higher rate (e.g., $rate=1.3$) accelerates the rising factor’s growth in later training stages, potentially leading to oscillations near the optimal solution and unstable convergence.
Therefore, setting the $rate=0.9$ achieves a balance between stability, smoothness, and sensitivity, making it an optimal choice in practice.

\subsubsection{Transferability and discriminability}

\begin{figure}[!t]
\centering
\includegraphics[width=0.7\linewidth,trim=255 150 255 150,clip]{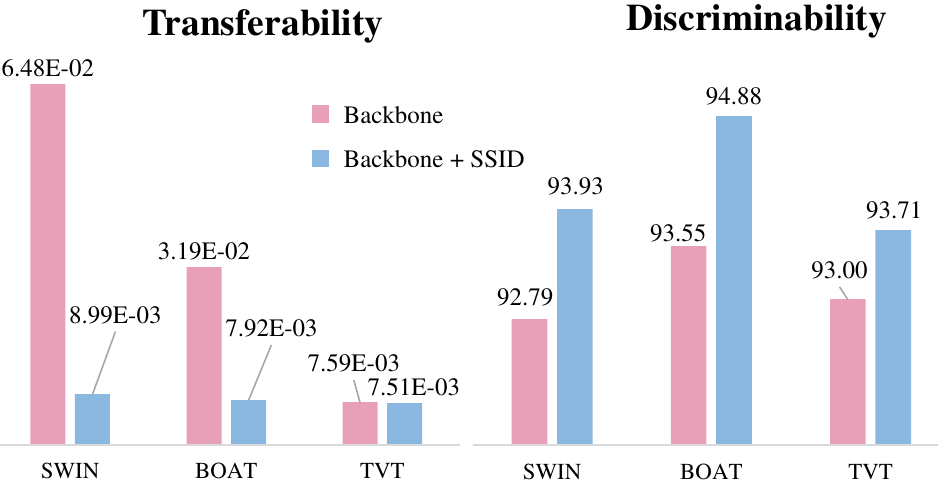}
\caption{\textbf{Left:} The transferability comparison: average inter-domain distance \({D_{s \leftrightarrow t}}\) of different backbones and backbone + SSID on VisDA-2017. {Right:} The discriminability comparison: accuracy of all samples in the source and target domains of different backbones and backbone + SSID on VisDA-2017 dataset.}
\label{fig_4}
\end{figure}

\begin{figure*}[!t]
\centering
\includegraphics[width=1\linewidth,trim=5 150 4 150,clip]{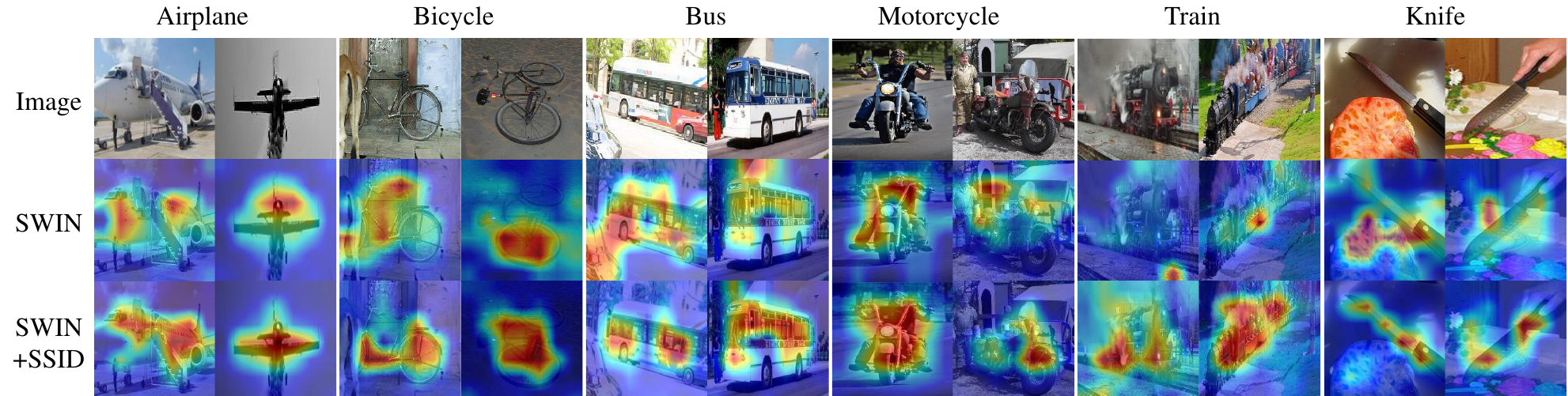}
\caption{Grad-CAM visualizations on the VisDA-2017. The "Image" rows display random original images, while the "SWIN" and "SWIN+SSID" rows show the Grad-CAM visualization results obtained by SWIN and SWIN+SSID, respectively. The columns correspond to different categories in the dataset, such as Airplanes, Bicycles, Bus, and so on. The network's attention is indicated by the intensity of red, with higher attention in the red regions and lower attention in the blue-violet regions.}
\label{fig_6}
\end{figure*}

Transferability refers to the ability of the classifier to transfer features from the source distribution to the target distribution, while discriminability refers to its ability to discriminate class information. As the target domain is completely unlabeled in UDA, improving both transferability and discriminability simultaneously is challenging. In this section, we investigate whether our proposed method can enhance the transferability of cross-domain features while implicitly preserving class-discriminative information for improved discriminability. 

After training, we use SSID with frozen model parameters to process all data and obtain the feature vectors \(F_s^{cls} \in {\mathbb{R}^{{N_s} \times {C^{cls}}}}\)  from the source domain and \(F_t^{cls} \in {\mathbb{R}^{{N_t} \times {C^{cls}}}}\) from the target domain before the classifier head layer on the VisDA-2017.
We then calculate their class center vectors \(\overline {F_s^{cls}} \in {\mathbb{R}^{K \times {C^{cls}}}}\) and \(\overline {F_t^{cls}} \in {\mathbb{R}^{K \times {C^{cls}}}}\) respectively, where \(K \) represents the number of classes. The average mean square error \((MSE)\) of each class center is calculated to represent the average inter-domain distance \({D_{s \leftrightarrow t}}\) between the source and the target feature distributions, as:
\begin{equation}
\label{deqn_ex1a}
{D_{s \leftrightarrow t}} = MSE(\overline {F_s^{cls}},\overline {F_t^{cls}}).
\end{equation}

The smaller distance indicates better transferability. Fig.~\ref{fig_4} illustrates the average inter-domain distance \({D_{s \leftrightarrow t}}\) for different backbones and backbone + SSID. 
% It's evident that SSID consistently reduces the distance between the source and target feature distributions. Hence, SSID is beneficial for enhancing transferability.
Notably, SSID consistently reduces the distance between source and target feature distributions, facilitating effective style transfer and improving target domain recognition without requiring manual annotation. Thus, SSID proves advantageous for enhancing transferability.

To further demonstrate the improvement of the proposed SSID in discriminability, we calculate the average accuracy of all samples in the source and target domains of the VisDA-2017 dataset to verify how well the class-discriminative information is preserved during the transfer process, as shown in Fig.~\ref{fig_4}.

% The results show that the proposed SSID performs better, with the overall classification accuracy of SWIN+SSID, BOAT+SSID, and TVT+SSID improving by 1.16\%, 0.88\% and 1.11\% over the SWIN, BOAT and TVT backbones, respectively. This demonstrates that SSID is able to retain more class-discriminative information and improve discriminability.
The results demonstrate that the accuracy of the target domain after style transfer is higher, with the overall classification accuracy of SWIN+SSID, BOAT+SSID, and TVT+SSID improving by 1.16\%, 0.88\% and 1.11\% over the SWIN, BOAT and TVT backbones, respectively. This indicates that style transfer is effective, and SSID helps retain more class-discriminative information, thereby enhancing the model's discriminative ability.

\subsubsection{Visualization Results}

In addition, we further compare the Grad-CAM visualizations~\cite{ref66} of SWIN and SWIN+SSID in each class on VisDA-2017 dataset, selecting six classes for presentation, as shown in Fig.~\ref{fig_6}. In the figure, red represents areas with high network attention, and blue-violet represents areas with low attention. 

It is evident that SWIN+SSID pays more attention to the discriminative features of objects, illustrating that SSID can improve the network's discriminability. Specifically, SWIN+SSID focuses on backbone skeleton features that reflect the object category, such as the head and wing of an airplane, the two tires and frame of a bicycle, the body of a train, and the blade of a knife. In contrast, SWIN's attention is somewhat scattered, focusing on partial features like the head of an airplane, a tire or frame of a bicycle, a small part of the train's body, and the area around a knife. 

Therefore, SWIN+SSID is better able to infer the correct class of objects based on regions of interest. Furthermore, when multiple classes of objects are present, SWIN+SSID can better focus on the correct class with fewer distractions. For example, in bus and motorcycle identification, SWIN focuses mostly on surrounding vehicles and cyclists, while SWIN+SSID focuses on the main body of the bus and motorcycle, paying almost no attention to surrounding objects.

These visualization results further demonstrate that SSID can enhance the discriminability of the network.

\section{Discussion and Conclusion}
In this study, we introduce a novel unsupervised domain adaptation (UDA) solution called Style-aware Self-Intermediate Domain (SSID). The proposed approach offers several distinct advantages: 1) SSID generates labeled, style-rich auxiliary features that effectively bridge domain gaps and facilitate knowledge transfer while preserving critical category-discriminating information. 2) External memory banks is leveraged to store and update the latest features, enabling the extraction of stable class-specific and category-style features for enhanced adaptability. 3) The innovative intra-domain and inter-domain loss functions are designed to simulate diverse latent feature spaces through unlimited sampling, enriching the model's generalization capabilities. 4) SSID is characterized by its plug-and-play design, making it easily integrable with various backbone networks, offering flexibility and scalability in practical applications. This combination of features positions SSID as a robust and versatile framework for addressing the challenges of domain adaptation.

However, we acknowledge the limitations of this study and recognize the need for further improvements. First, the explicit involvement of intermediate domain images in the training process incurs memory overhead. Future studies should focus on developing implicit intermediate domains that eliminate the need for additional memory resources. Additionally, we assume each class domain follows a normal distribution, which may not always hold in practice and could affect the method's effectiveness. It would be worthwhile to explore alternative strategies to accommodate non-Gaussian distributions, such as leveraging more flexible distribution models or employing non-parametric techniques.

Moving forward, we aim to extend our exploration to other data types, such as speech and text, and investigate adaptive methods for handling multiple modalities. Additionally, the generalization capability of the proposed approach will be validated on larger and more diverse datasets. Investigating its adaptability to tasks like segmentation and object detection also presents an exciting avenue for future research. Furthermore, we will explore strategies for generating high-quality intermediate domains to achieve robust domain generalization in scenarios where the target domain is unknown.

In conclusion, we propose a novel Style-aware Self-Intermediate Domain (SSID) to bridge significant domain gaps and transfer knowledge while mitigating the loss of class-discriminative information. We conduct comprehensive experiments on widely used domain adaptation benchmarks to evaluate the effectiveness of the proposed SSID. The results demonstrate that SSID seamlessly integrates with various backbone networks as a plug-and-play module, delivering enhanced performance. This method holds the potential to reduce diagnostic errors and provide a reliable solution for real-world applications, paving the way for broader and more impactful advancements in domain adaptation research.

\section*{Acknowledgments}
This work was supported by the National Natural Science Foundation of China (Nos. 62136004, 62276130), the Key Research and Development Plan of Jiangsu Province (No. BE2022842), and Huazhu Fu's A*STAR Central Research Fund and Career Development Fund (C222812010).

\appendix
\section{Algorithm of style-aware Self-Intermediate Domain (SSID)}
\label{app1}
We summarize the strategy of our proposed SSID in Algorithm~\ref{alg:alg1}

\begin{algorithm}
\caption{Style-aware Self-Intermediate Domain (SSID).}\label{alg:alg1}
\begin{algorithmic}[1]
\REQUIRE{The source domain \(S\), SSID \(I\), the target domain \(T\), number of feature extractor blocks \(L\), and batch size \(B\).}
\STATE Initialize \(I\) with \(S\). 
\STATE Initialize the memory bank.
\STATE \textbf{For \(epoch=1\) to \(Maxepochs\) do}
\STATE \textbf{\qquad For \(l=1\) to \(L\) do}
\STATE \qquad \qquad Calculate the output of \(x_s\), \(x_i\), \(x_t\) in the \(l\)-th feature extractor block: \(f_s^l\), \(f_i^l\), \\
$ \text{\qquad \qquad and \(f_t^l \).} $
\STATE \qquad \qquad Randomly select \(B/3\) features from \(f_s^l\), \(f_i^l\) and \(f_t^l \) to form \(f_{sample}^l\).
\STATE \qquad \qquad Update \(f_i^l\) according to (\ref{eq2}).
\STATE $ \textbf{\qquad End For} $
\STATE \qquad update $\left\{ {M_\mu ^{{N_s}},M_\sigma ^{{N_s}},M_{cls}^{{N_s}},M_\mu ^K,M_\sigma ^K,M_{cls}^K} \right\}$ .
\STATE \qquad Update framework with the intra-domain loss function (\ref{eq5}),~(\ref{eq6}) and (\ref{eq7}).
\STATE \qquad Update framework with the inter-domain loss function (\ref{eq9}) and (\ref{eq10}).
\STATE \textbf{End For}
\end{algorithmic}
\end{algorithm}

\bibliographystyle{elsarticle-num}
\bibliography{ref} 
\end{document}